\documentclass[sigconf]{acmart}

\AtBeginDocument{%
  }

\setcopyright{acmlicensed}
\copyrightyear{2018}
\acmYear{2018}
\acmDOI{XXXXXXX.XXXXXXX}
\acmConference[Conference acronym 'XX]{Make sure to enter the correct
  conference title from your rights confirmation email}{June 03--05,
  2018}{Woodstock, NY}
\acmISBN{978-1-4503-XXXX-X/2018/06}

\usepackage[utf8]{inputenc} 
\usepackage[T1]{fontenc}    
\usepackage{hyperref}       
\usepackage{url}            
\usepackage{booktabs}       
\usepackage{amsfonts}       
\usepackage{nicefrac}       
\usepackage{microtype}      
\usepackage{xcolor}         
\usepackage{makecell}
\usepackage{amsthm} 
\usepackage{multirow}

\usepackage{amsmath}
\usepackage{mathtools}
\usepackage{amsthm}
\usepackage{diagbox}
\usepackage{tabularx}

\usepackage{bm}
\usepackage{graphicx}
\usepackage{tikz}
\usetikzlibrary{positioning,arrows.meta,calc,shapes.misc,fit}
\usepackage{multirow}
\usepackage{colortbl}
\usepackage{rotating}
\usepackage{makecell}
\usepackage{wrapfig}
\usepackage{enumitem}
\usepackage{pifont}
\usepackage{makecell}
\usepackage[ruled,vlined,linesnumbered]{algorithm2e}

\usepackage[most]{tcolorbox}

\usepackage{multicol}
\usepackage{setspace}

\definecolor{best}{rgb}{1,  0,  0}
\definecolor{second}{rgb}{0,  0,  0.859}
\definecolor{yellowhl}{rgb}{1, 0.97, 0.7}
\newcommand{\best}[1]{\textcolor{best}{\underline{#1}}}
\newcommand{\second}[1]{\textcolor{second}{\textbf{#1}}}
\newcommand{\ybest}[1]{\colorbox{yellowhl}{#1}}
\usepackage{hyperref}
\SetCommentSty{footnotesize}
\SetKwComment{Comment}{$\triangleright$ }{}
\SetSideCommentRight

\newcommand{\method}{\textsc{Atlas}\xspace}
\theoremstyle{remark}

\tcolorboxenvironment{proposition}{
  notitle,
  rounded corners,
  colframe=darkgray,
  colback=blue!5,
  boxrule=0.75pt,
  boxsep=0pt,
  left=0.2cm,
  right=0.2cm,
  enhanced,
  shadow={1pt}{-1pt}{0pt}{opacity=0.5,gray},
  toprule=0.75pt,
  before skip=0.65em,
  after skip=0.75em
}
\tcolorboxenvironment{lemma}{
  notitle,
  rounded corners,
  colframe=darkgray,
  colback=blue!5,
  boxrule=0.75pt,
  boxsep=0pt,
  left=0.2cm,
  right=0.2cm,
  enhanced,
  shadow={1pt}{-1pt}{0pt}{opacity=0.5,gray},
  toprule=0.75pt,
  before skip=0.65em,
  after skip=0.75em
}
\tcolorboxenvironment{corollary}{
  enhanced, breakable,
  colback=gray!6, colframe=gray!55,
  boxrule=0.4pt, arc=1.5pt,
  left=4pt, right=4pt, top=3pt, bottom=3pt,
  before skip=6pt, after skip=6pt,
}
\tcolorboxenvironment{remark}{
  enhanced, breakable,
  colback=gray!6, colframe=gray!55,
  boxrule=0.4pt, arc=1.5pt,
  left=4pt, right=4pt, top=3pt, bottom=3pt,
  before skip=6pt, after skip=6pt,
}

\renewcommand\footnotetextcopyrightpermission[1]{}

\begin{document}

\title{Every Client Is an Environment: Federated De-confounding for Spatio-Temporal Forecasting}


\author{Qingxiang Liu\textsuperscript{\rm 1}, Anqi Liang\textsuperscript{\rm 2},  Heng Wang\textsuperscript{\rm 1,3}, Yuxuan Liang\textsuperscript{\rm 1,}} 
\authornote{Corresponding Author.}
\affiliation{%
  \institution{
  \textsuperscript{\rm 1}The Hong Kong University of Science and Technology (Guangzhou) \hspace{0.1em} \\
  \textsuperscript{\rm 2} Shanghai Jiao Tong University,
  \textsuperscript{\rm 3} Harbin Engineering University\\
  }
  \city{} 
  \state{}
  \country{}
}

\email{qingxiangliu737@gmail.com, lianganqi@sjtu.edu.cn}
\email{heng.wang@hrbeu.edu.cn, yuxliang@outlook.com}

\renewcommand{\shortauthors}{Liu et al.}

\begin{abstract}
Federated learning has emerged as a promising paradigm for spatio-temporal forecasting (STF), enabling collaborative model training without sharing raw observations.
Existing federated STF methods primarily regard cross-client heterogeneity as an optimization challenge and mitigate it through personalized approaches. 
However, such heterogeneity fundamentally stems from diverse \emph{environmental conditions}, and these methods capture environment-specific forecasting patterns, hardly generalizing under environmental shifts.
Our key insight is that the environmental diversity across federated clients should be exploited, as they provide \emph{complementary observations of the same underlying spatio-temporal system}.
Based on this insight, we propose \method, a novel federated de-confounding framework that \textbf{treats clients as distinct causal environments}. 
\method leverages the client heterogeneity as distributed environmental evidence and learns a global prototype codebook to capture shared environmental regimes.
We further derive a theoretical federated de-confounding bound that is linearly controlled by the averaged confounding strength. 
Extensive experiments demonstrate that \method consistently outperforms federated baselines, while providing transferable, interpretable, and communication-efficient environmental representations.
\end{abstract}



\received{20 February 2007}
\received[revised]{12 March 2009}
\received[accepted]{5 June 2009}

\maketitle

\begin{figure}
    \centering
    \includegraphics[width=\linewidth]{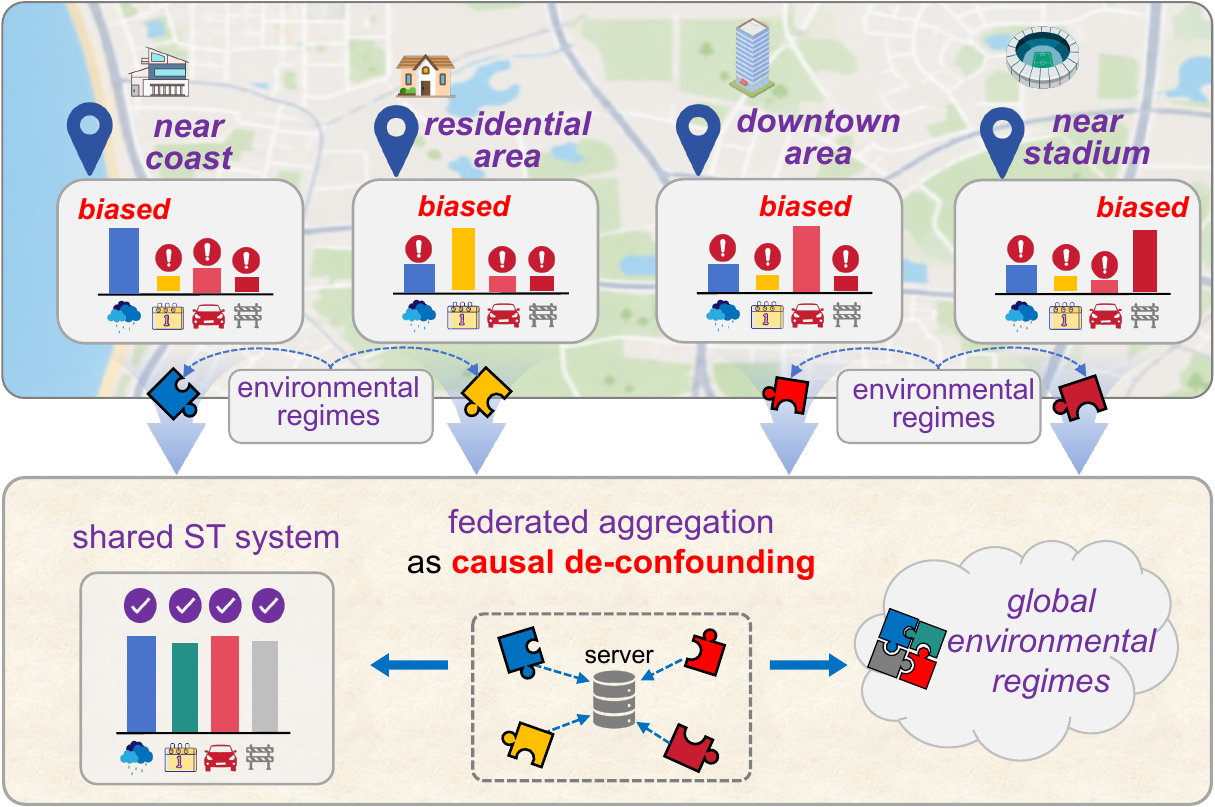}
    \caption{Clients are treated as environments, and federated aggregation promises the shared environmental regimes to eliminate client-specific confounding effects.}
    \label{fig:intro-insight}
\end{figure}

\section{Introduction}
\label{sec:intro}
Spatio-temporal forecasting (STF) plays a pivotal role in urban computing by predicting traffic trends, air quality, and service demand to empower intelligent transportation, environmental monitoring, and public-health decisions~\citep{fang2026unraveling,mao2025survey,zhang2024urban}.
Recent studies have achieved remarkable performance by training deep-learning models based on historical observations~\citep{li2017dcrnn,wu2019graphwavenet,xia2023cast,zhao2026fast}. 
However, these methods generally assume that all training data can be accessed and used collectively for model optimization. 
In real-world scenarios, the raw time series readings are from distributed traffic nodes or edge devices deployed by different operators, which cannot be uploaded to a centralized server due to strict regulations or communication constraints. 
Federated learning (FL) has emerged as a promising approach to tackle this limitation by sharing intermediate model parameters, rather than raw time series data, among different nodes (termed \emph{clients} in FL)~\citep{mcmahan2017fedavg,liu2020fedgru,fuels2025,meng2021cnfgnn}.

In the federated STF paradigm, the time series observations on each client have heterogeneous distributions due to varying deployment locations.
Existing federated STF methods have tried to tackle this challenge using various personalized strategies~\citep{feddis,yang2024fedgtp,SCFSGL}.
For example, Liu et al.~\citep{fuels2025} leverage contrastive learning to enlarge the diversity of cross-client models.
Zhang et al.~\citep{zhang2024modeling} introduce adaptive spatial prompts to augment local information.
Despite the performance gains, these methods mostly regard the observable heterogeneity as the optimization challenge to be tackled, rather than investigating \emph{the underlying reasons of environmental effects}.

As shown in Figure~\ref{fig:intro-insight}, each traffic node observes spatio-temporal dynamics within its local environmental context, such as meteorological factors, special events, and rush-hour patterns.
The limited environmental exposure causes local models to capture environment-specific forecasting patterns, where the intrinsic forecasting mechanism is entangled with client-specific confounding effects.
Therefore, these local models fail to generalize when environmental conditions shift among locations or over time.
However, the diverse environments across clients provide complementary perspectives of the shared spatio-temporal system, \textit{making it possible to distangle environment-conditioned confounding effects}.
Motivated by this, we do not treat cross-client heterogeneity as a nuisance to be suppressed, but rather \textbf{consider federated clients as naturally distributed causal environments}.

In this paper, we revisit the cross-client heterogeneity in federated STF from the perspective of causal de-confounding.
Primarily, we formulate the \emph{Structural Causal Model} (SCM) for federated STF to characterize how environmental conditions induce client-specific spurious correlations between historical observations and future horizons~\citep{pearl2009causality}. 
Based on SCM, we then analyze why an individual client cannot eliminate confounding effects via \emph{back-door adjustment}.
We propose \textbf{\method}, a novel federated de-confounding framework, where a shared prototype codebook is adopted to capture the common regularities of environmental effects, thereby reducing client-specific confounding. 
Finally, we theoretically justify the proposed \method and extensively validate its effectiveness and efficiency through empirical studies.

The contributions of our paper are summarized as follows.
\begin{itemize}[leftmargin=*]
\item \textbf{A federated de-confounding framework.} We introduce a causal perspective that considers each federated clients as casual environments.
Based on this insight, we propose \method to learn the shared environmental confounding effects among heterogeneous clients.
\item \textbf{Federation as an implicit back-door adjustment.} 
We theoretically reveal that federated codebook aggregation establishes the conditions for mitigating environmental confounding, thereby acting as an implicit back-door adjustment.
We derive a federated de-confounding error bound, which demonstrates that the error is linearly controlled by the averaged confounding strengths, rather than each individual client.
\item \textbf{Empirical performance.} 
Extensive experiments are conducted on five real-world datasets to validate the effectiveness, efficiency, transferability, and interpretability of our proposed \method.
\end{itemize}

\section{Related Works}

\textbf{Federated Spatio-Temporal Forecasting.}
In recent years, spatio-temporal forecasting has been studied extensively. Existing methods typically model the temporal correlation using recurrent neural networks~\citep{graves2013generating} or temporal convolutional networks~\citep{bai2018empirical}, while capturing the spatial correlation using graph neural networks~\citep{kipf2016semi}. Subsequent studies have further improved forecasting performance by incorporating graph attention mechanisms~\citep{guo2019attention,zheng2020gman}, Transformer architectures~\citep{liang2023airformer}, and mixture-of-experts models~\citep{jacobs1991adaptive,zhao2026fast}.
Despite their promising performance, these methods rely on centrally collected data. In practice, spatio-temporal data are often distributed across regions, organizations, or devices and cannot be centrally collected due to ownership, communication, or privacy constraints. Federated learning offers a suitable paradigm for such decentralized settings, allowing participants to collaboratively train a shared forecasting model without sharing raw data.
Early methods such as FedGRU~\cite{liu2020fedgru} directly combine federated averaging with recurrent forecasting models.
CNFGNN~\cite{meng2021cnfgnn} separates temporal modeling in clients from spatial modeling on the server. 
More recent studies have focused on addressing cross-client heterogeneity. 
FedGTP~\cite{yang2024fedgtp} recovers inter-client spatial dependencies under privacy constraints.
Fuels~\cite{fuels2025}, FedDis~\cite{feddis}, and SC-FSGL~\cite{SCFSGL} aim to decouple global and personalized patterns using contrastive learning.
Despite these advances, existing methods still treat inter-client heterogeneity as an optimization challenge to be mitigated, leaving forecasting performance vulnerable to underlying environmental shifts.

\noindent\textbf{Causal Spatio-Temporal Forecasting.}
Causal inference provides a principled framework for distinguishing invariant mechanisms from spurious correlations through techniques such as back-door adjustment and invariant prediction~\citep{pearl2009causality,peters2016icp,arjovsky2019irm}. These approaches have recently been applied to STF.
Specifically, CaST~\citep{xia2023cast} uses back-door adjustment to separate invariant patterns from temporal environments and front-door adjustment to capture spatial ripple effects.
STEVE~\citep{ji2025seeing} extends back-door adjustment to unknown confounders by learning a bank of basis confounder representations.
CauSTG~\citep{zhou2023caustg} partitions temporal steps into sub-environments and identifies spatio-temporal relations that remain invariant across them. 
STONE~\citep{wang2024stone} and subsequent studies~\citep{ma2025robust,yang2025revealing} construct or infer diverse environments to address spatio-temporal distribution shifts.
EAGLE~\citep{yuan2023eagle} disentangles latent environments and identifies invariant patterns through node-wise causal interventions.
NuwaDynamics~\citep{wang2024nuwadynamics} and CaPaint~\citep{capaint2024} identify causal patches and intervene on non-causal patches using spatio-temporal mixup and diffusion-based inpainting, respectively.
However, in a federated setting, diverse environmental conditions are distributed across clients, while the lack of centralized access to raw data makes it difficult to disentangle environmental confounding effects.

\noindent\textbf{Causal Learning Across Federated Clients.}
A parallel line of work studies causal learning for out-of-distribution (OOD) generalization in federated settings on other tasks~\cite{li2025federated}.
FedSDR~\cite{tang2024learning} and FedPIN~\cite{tang2024causally} distinguish personalized features from spurious features in object detection.
FedDDL~\cite{qi2025federated} performs federated back-door adjustment to remove confounding paths during model inference for image classification.
FedCIFL~\cite{guo2025federated} employs causal client selection to mitigate spurious correlations in sentiment classification.
However, federated STF presents a distinct challenge, where environmental confounders are latent and distributed across clients, affecting forecasting through complex spatial and temporal dependencies.
Consequently, existing methods cannot effectively address spatio-temporal heterogeneity or cross-client causal confounding.

\section{Formulation}

We denote each given node $k (1\leq k\leq N)$ as a client in the federated paradigm.
Let $x_k^t\in\mathbb{R}^d$ denote the readings with $d$ features at time step $t$.
For each client, given the historical readings from the preceding $T$ time steps, $X_k^{(t-T+1):t}\in\mathbb{R}^{T\times d}$,
each client aims to predict the future $H$ steps, $Y_k^{(t+1):t+H}\in\mathbb{R}^{H\times d}$.
For simplicity, we refer to $X_k^{(t-T):t}$ and $Y_k^{(t+1):t+H}$ as $X_k$ and $Y_k$ respectively.
Different from the conventional STF methods, we omit the topology graph across nodes here.
The spatial correlation can be implicitly learned by aggregation in the server.

We aim to learn the personalized prediction model $f_{\theta_k}(\cdot)$ for each client $k$ in a federated manner, where $\theta_k$ denotes the model parameters.
Therefore, the target of federated spatio-temporal forecasting is formulated as:
\begin{equation}
  \min_{\{\theta_k\}} \sum_k \sum_{(X_k, Y_k)\in \mathcal{D}_k} \mathcal{L}(f_{\theta_k}(X_k), Y_k),
\end{equation}
where $\mathcal{L}$ evaluates the forecasting errors, and $\mathcal{D}_k$ denotes the local dataset of client $k$.

\begin{figure}[!t]
\includegraphics[width=\linewidth]{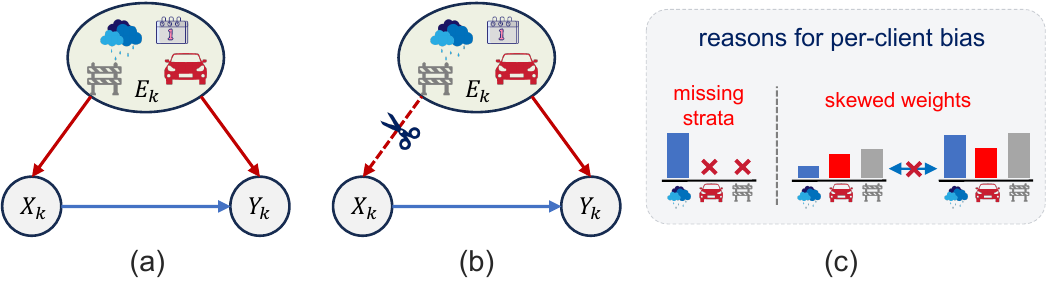}
\caption{Causal analysis of federated STF, with (a) client-specific SCM, (b) back-door adjustment, and (c) reasons for per-client de-confounding bias.}
\label{fig:avg_intut}
\end{figure}

\noindent\textbf{Local Structural Causal Model.}
From the causal perspective, we can construct a structural causal model (SCM) to understand the underlying generation mechanisms of spatio-temporal data at each client~\citep{pearl2009causality}. 
The SCM includes three components, i.e., the historical observation $X_k$, the future horizon $Y_k$, and the client-specific environmental conditions $E_k$.
As shown in Figure~\ref{fig:avg_intut}(a), $E_k$ introduces spurious correlations between the input and target through the back-door path $X_k\leftarrow E_k \rightarrow Y_k$.
For example, environmental conditions like weather and events affect the temporal distribution.

\noindent\textbf{De-confounding via Back-door Adjustment.}
In causal analysis, the {back-door adjustment} is an approach used to estimate the causal effect by adjusting confounding variables~\cite{xia2023cast,sui2022causal}. 
The \textit{do-calculus} operator is adopted to estimate $P(Y_k|do(X_k))$ so as to cut the back-door path $E_k \rightarrow X_k$, the red dashed line in Figure~\ref{fig:avg_intut}(b).
We provide the details of back-door adjustment in Sec.~\ref{sec:why-fail}.
The confounding correlation can be eliminated and the causal effect from $X_k$ to $Y_k$ can be obtained.

\section{Methodology}
\label{sec:method}
In this section, we firstly analyze why intra-client adjustment cannot eliminate environmental confounding effects. 
We then revisit the federated aggregation as the de-confounding process based on the transferred prototype codebook.
The federated de-confounding errors are validated by both theoretical analysis and empirical study.
We finally propose the \method as an implementation of the federated de-confounding framework.

\subsection{Why Local Clients Cannot De-confound}
\label{sec:why-fail}
The do-calculus operator is adopted to estimate $P(Y\mid do(X))$ so as to block the back-door path~\citep{xia2023cast} by stratifying temporal environment into $I$ discrete strata
$E=\{e^1,\dots,e^I\}$ and marginalizing over them:
\begin{equation}
  P(Y|do(X)) = \sum_{i=1}^I P(Y|X, E{=}e^i)\,P(E{=}e^i).
  \label{eq:backdoor}
\end{equation}
Since the centralized setting has access to the complete marginal distribution of $E$, i.e., $P(E)$, the back-door treatment is promised by estimating every stratum $P(Y|X,E{=}e^i)$.
As we analyzed in Sec.~\ref{sec:intro}, in the federated setting, raw data are kept locally, and no individual client observes the full set of environmental strata.

Before dedicated analysis, we define the two predictors:
\begin{definition}[\textbf{Local estimate}]
\label{def:local-estimate}
The \emph{local estimate} $b_k$ is denoted as $ b_k=\sum_{e} P(Y_k\mid X_k,E=e)\,P_k(E=e)$, which is the adjustment a client can conduct from its own data.
The strata are weighted on the local marginal distribution $P_k(E)$. 
We assume that all clients are located in a shared spatio-temporal dynamics system.
Therefore, under the same environmental conditions, the generated spatio-temporal observations are homogeneous, i,e, shared $P(Y_k\mid X_k,E)$.
\end{definition}
\begin{definition}[\textbf{De-confounding target}]
\label{def:deconf-target}
The \emph{de-confounding target} $\tau$ is denoted as $\tau=\sum_e P(Y_k\mid X_k,E=e)\,P(E=e)$, which averages the environmental strata on the weights of the global marginal distribution $P(E)$.
\end{definition}
We therefore can obtain the \emph{de-confounding bias} as
\begin{equation}
\begin{split}
  \Delta_k = b_k-\tau= \sum_e P(Y_k\mid X_k,e)\,\big(P_k(e)-P(e)\big).
\end{split}
  \label{eq:node-bias}
\end{equation}

We analyze that such gap results from the following two aspects and neither of them can be tackled by individual client, but rather by restoring the global environmental distribution through federation.
\begin{itemize}[leftmargin=*]
  \item[(i)] \textbf{Missing strata.} 
    The conditional distribution between historical observations and future horizons varies across environmental strata. 
    Each client observes only a biased subset of these strata.
  As shown in Figure~\ref{fig:avg_intut}(c)(left), for a traffic node deployed at a location where the traffic flow is primarily influenced by weather conditions, the local environmental distribution $P_k(E)$ is concentrated on weather-related strata, providing limited coverage of other strata associated with special events or rush-hour conditions.
  Federation tackles this limitation by aggregating diverse environmental strata observed across heterogeneous clients.
  \item[(ii)] \textbf{Skewed weights.} The local estimate is weighted according to the local environmental distribution $P_k(E)$ rather than the global distribution $P(E)$. As shown in Figure~\ref{fig:avg_intut}(c)(right), the local and global environmental distributions assign different weights to the same strata, leading to a biased local estimate.
  Federation calibrates the environmental distribution by aggregating complementary environmental strata across clients, thereby providing an unbiased estimate of the environmental confounding effects.
\end{itemize}

\subsection{How Federation Makes De-confounding}
\label{sec:proto}
\label{sec:setup}
Since clients encounter different environmental strata, a natural intuition is that their local environmental distributions can be aggregated to approximate the global distribution, i.e., $P(E)\leftarrow\sum_k w_k P_k(E)$.
Here, $w_k$ denotes the aggregation weights.
We can derive that aggregating local back-door estimates approximates the global adjustment $\tau$ and mitigates client-specific bias $\Delta_k$:
\begin{equation}
  \begin{split}
    \tau(X_k) &=\sum_e P(Y_k\mid X_k,e)\,P(e) \\
  & \overset{\triangle}{=} \sum_e P(Y_k\mid X_k,e)\sum_k w_k P_k(e) = \sum_k w_k\,b_k(X_k).
  \end{split}
  \label{eq:fed-debias}
\end{equation}

From the perspective of representation learning, we introduce a compact codebook $Q=[q^1,\dots,q^I]$, with prototype $q^i\in\mathbb{R}^{D}$ corresponding to an environmental stratum.
The client $k$ encodes its input with a personalized backbone, $z_k=\varphi_k(X_k)\in\mathbb{R}^{D}$.
We obtain the environment-conditioned representation $\hat z_k \in \mathbb{R}^D$ by softly assigning $z_k$ to the prototype codebook:
\begin{equation}
  \hat z_k = A(z_k;Q_k) = \operatorname{softmax}\!\big(\cos(z_k,Q_k^\top)/\alpha\big)\,Q_k,
  \label{eq:assign}
\end{equation}
where temperature $\alpha>0$ is a temperature parameter controlling the sharpness of prototype assignment. A larger $\alpha$ yields smoother mixtures over environmental regimes.

\begin{tcolorbox}[
  notitle,
  rounded corners,
  colframe=darkgray,
  colback=blue!5,
  boxrule=0.75pt,
  boxsep=0pt,
  left=0.2cm,
  right=0.2cm,
  enhanced,
  shadow={1pt}{-1pt}{0pt}{opacity=0.5,gray},
  toprule=0.75pt,
  before skip=0.65em,
  after skip=0.75em
]
\textbf{\textsc{Lemma} 1 (Prototype-Induced Approximation of Local Adjustment).}
A client's codebook $Q_k=[q_k^1,\dots,q_k^I]$ partitions the representation space into $I$ discrete components.
Each prototype $q_k^i$ encodes the forecasting mechanism associated with environmental stratum $e^i$ estimated from local observations. 
The soft assignment in Eq.~\eqref{eq:assign} reorganizes $\hat z_k$ as a convex combination of these prototype vectors. 
Therefore, $\hat z_k$ functions as a parameter-level approximation of the local back-door adjustment $b_k$ and  inherits the de-confounding bias $\Delta_k$ defined in Eq.~\eqref{eq:node-bias}.
\end{tcolorbox}

\noindent\textbf{Alignment and Aggregation.}
\label{sec:agg}
$\hat z_k$ is computed locally on each client and only the codebook $Q_k$ is communicated in federated optimization.
The prototype slots are exchangeable. 
For example, the $i$-th slots in client $k$ and $j$ may specialize to different environmental regimes.
Therefore, simply averaging by slot index may perturb the codebook space.
The server has to align prototypes across clients firstly. 
In each round, the server evaluates every $(anchor, client)$ slot pair by cosine distance:
\begin{equation}
  C_k^{i,j} = 1 - \cos\!\big(q^i,\; q^j_k\big)\in[0,2],
  \quad i,j\in\{1,\dots,I\},
  \label{eq:align-cost}
\end{equation}
where $q^i\in Q$ (the global codebook) is anchor slot $i$ and $q^j_k$ is client $k$'s slot $j$.
We denote $\bm{C}_k =\{\,C_k^{i,j} \mid 1\leq i,j\leq I\,\} \in \mathbb{R}^{I\times I}$.
We seek the optimal permutation $\Pi_k$ of minimal total cost
\begin{equation}
  \Pi_k =\arg\min_{\Pi\in\mathcal{C}}\ \langle \Pi,\,\bm{C}_k\rangle
          =\arg\min_{\Pi\in\mathcal{C}}\ \sum_{i,j}\Pi^{i,j}\,C_k^{i,j},
  \label{eq:assign-prob}
\end{equation}
where $\mathcal{C}$ is the set of $I\times I$ permutation matrices, and each entry $\Pi^{i,j}\in\{0,1\}$.
$\Pi^{i,j}{=}1$ means client $k$'s slot $j$ is matched to anchor slot $i$. 
Every client's slot is mapped with exactly one anchor slot. 
The Eq.~\eqref{eq:assign-prob} is a linear assignment problem of finding the matching with minimal total cost, which can be solved exactly in $O(I^3)$ by the Hungarian algorithm~\citep{kuhn1955hungarian}.
The aligned codebook $\Pi_kQ_k$ denotes that the client $k$'s rows are  reordered into the anchor's indexing. 

The server can perform the weighted aggregation:
\begin{equation}
  Q \;\leftarrow\; \textstyle\sum_k w_k\,\Pi_k Q_k,
  \label{eq:agg}
\end{equation}
where $w_k=|\mathcal{D}_k|/\sum_k |\mathcal{D}_k|$ is proportional to the data size.
We then theoretically analyze why the aggregated codebook $Q$ recovers the global environmental distribution $P(E)$ and promises de-confounding.

\noindent\textbf{Federation Mitigates Environmental Confounding.}
Each local prototype codebook is learned under the client's environmental distribution $P_k(E)$, which can introduce client-specific bias from the theoretically optimal codebook induced by the global distribution $P(E)$.
Federated learning aggregates these local codebooks and fulfills the approximation of $\sum_k w_kP_k(E)$.
Therefore, federation recovers the global environmental distribution $P(E)$ and eliminates the client-specific confounding.

We theoretically characterize this de-confounding bound in \textsc{Proposition}~1, with the detailed proof in Appendix~\ref{sec:prop}.
The proposition shows that the federated de-confounding error is linearly dependent on the averaged confounding strengths.

\begin{tcolorbox}[
  notitle,
  rounded corners,
  colframe=darkgray,
  colback=blue!5,
  boxrule=0.75pt,
  boxsep=0pt,
  left=0.2cm,
  right=0.2cm,
  enhanced,
  shadow={1pt}{-1pt}{0pt}{opacity=0.5,gray},
  toprule=0.75pt,
  before skip=0.65em,
  after skip=0.75em
]
\textbf{\textsc{Proposition} 1 (Federated De-confounding Bound).}
\label{prop:deconfound}
After federated codebook aggregation, the federated de-confounding bias is
bounded by

\begin{equation}
\left\|
\hat z-z^*
\right\|
\leq
O\left(
\left\|
\sum_k w_k\Delta_k^Q
\right\|
\right),
\label{eq:deconf-error}
\end{equation}
where $z^*$ denotes the representation conditioned on the optimal codebook $Q^*$ and $\Delta_k^Q$ denotes the client-specific prototype deviation induced
by local environmental confounding. 
Therefore, the de-confounding error is \emph{linearly} dependent on averaged confounding strength, manifested as prototype deviation.
Detailed proof is provided in Appendix~\ref{sec:prop}.
\end{tcolorbox}

\begin{figure}[!t]
  \centering
  \includegraphics[width=\linewidth]{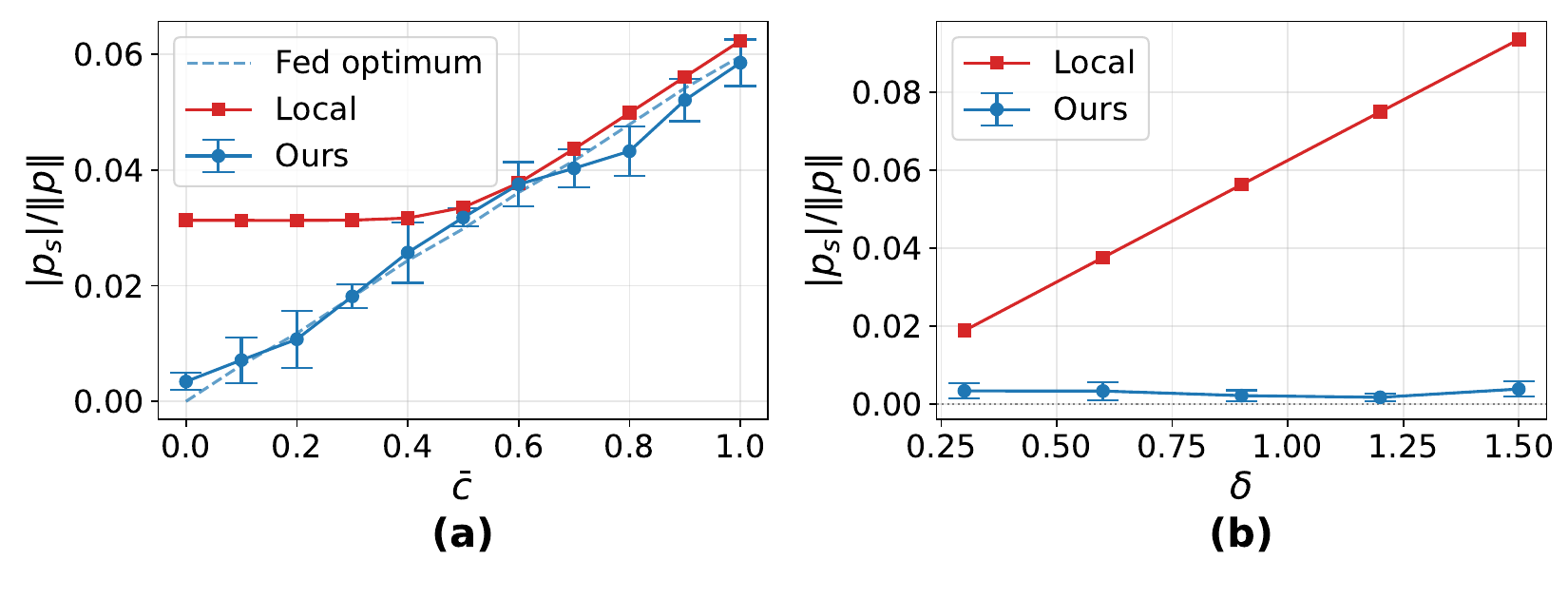}
  \caption{Empirical validation of \textsc{Proposition}~1. The federated de-confounding error is approximately linearly with averaged confounding strengths.}
  \label{fig:inv}
\end{figure}

\subsection{Empirical Validation}
\label{sec:empirical-validation}
\textsc{Proposition}~1 demonstrates that the federated de-confounding bound is linearly dependent on the averaged confounding strength and independent of individual client.
In this subsection, we validate the proposition under a semi-synthetic setting to inject the controlled spurious correlation into the real METR-LA traffic data.

\noindent\textbf{Semi-synthetic Data.}
Let $x_k = \text{mean}(X_k)$ and $y_k = \text{mean}(Y_k)$ denote the input context and forecasting target.
A spurious variable is obtained as:
$s_k = c_k \cdot y + \xi$, where $\xi\sim\mathcal{N}(0,\sigma_\xi^2)$ is
independent noise with $\sigma_\xi{=}2.0$.
The scalar $c_k$ controls the confounding strength of the spurious feature $s_k$ with the target $y_k$.
We can obtain $c_k = \bar c + \delta\cdot r_k$, where $\bar c$ and $\delta$ are the two independent parameters and $r_k\in\{\pm1\}$ is a fixed sign pattern balanced across clients, with $\sum_k r_k=0$. We can always have $\sum_k w_k c_k = \bar c$, with $w_k = 1/N$.
Here, $\bar c$ represent the averaged confounding strength across all clients, and $\delta$ controls the magnitude of individual confounding effect.
The observed input is $u_k=[x_k,s_k]\in\mathbb{R}^2$. From the model's perspective, $x_k$ and $s_k$ are indistinguishable, reflecting the potential confounding factors prevalent in real-world time series.

\noindent\textbf{Settings.}
We adopt a single shared prototype $p\in\mathbb{R}^2(i.e., I=1)$.
The local forward process can be represented as $\hat{y}_k = a_k(u_kp_k^\top) + b_k$, where $a_k$ and $b_k$ are two scalars for optimization.
At each round, the local updates of $p_k(1\le k\le N)$ are averaged.
We introduce the \emph{confounding leakage}, which can be obtained as $\ell(p)=|p_s|/\|p\|\in[0,1]$, representing the magnitude of the spurious component $p_s$ (the second coordinate of $p$) relative to the learned prototype's norm. 
At $\ell(p){=}0$, the direction aligns purely with the input feature $x$, while it collapses
onto the spurious feature at $\ell(p){=}1$.
We simulate $N=30$ clients with $2000$ samples for $60$ federated rounds. 
We conduct 5 trials and report the mean and standard deviation values.
In \emph{Fed optimum} setting, we exhaustively search for the optimized shared $p$ under the same model architecture.
In \emph{local} setting, each client trains its specific prototype locally.

\noindent\textbf{Findings.}
Figure~\ref{fig:inv}(a) varies $\bar c\in[0,1]$ with fixed $\delta=0.5$. 
The confounding leakage $\ell(p)$ increases approximately linearly with $\bar c$ and passes through $(0,0)$, closely matching the Fed optimum. 
When $\bar c=0$, positive and negative client-specific confounding effects cancel exactly, allowing the shared prototype to recover the invariant axis ($\ell(p)\approx0$). 
In contrast, local setting exhibits consistently high leakage across all values of $\bar c$.
Figure~\ref{fig:inv}(b) provides a complementary control by fixing $\bar c=0$ and varying $\delta\in[0.3,1.5]$. As $\delta$ increases, confounding leakage increases in local setting because larger $|c_k|$ strengthens the spurious correlation within each client. 
However, the confounding leakage remains near zero in \method, indicating that it depends on the averaged confounding strength $\bar c$ rather than the magnitude of individual client's confounder.
These results demonstrate that the client-specific spurious correlation that biases local treatment can be effectively mitigated through federated de-confounding.

\subsection{Model Instantiation}
\label{sec:instantiation}

\begin{figure}[!t]
\includegraphics[width=\linewidth]{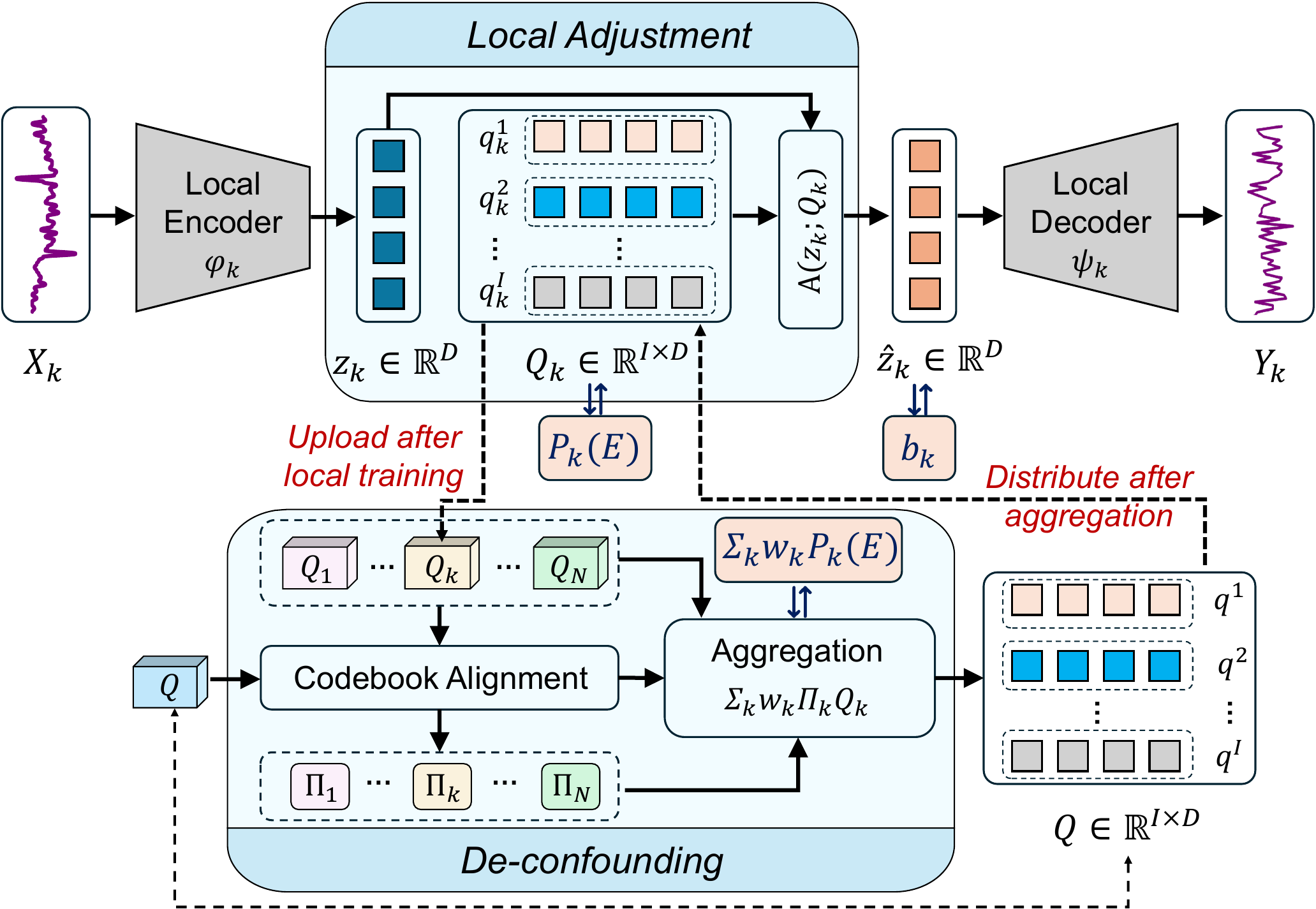}
\caption{The overall implementation of \method. Each component is designed exactly according to federated de-confounding mechanism.}
\label{fig:overview}
\end{figure}

\textbf{Model Architecture.}
We implement the federated de-confounding mechanism by proposing the \textbf{\method} framework, as shown in Figure~\ref{fig:overview}.
The encoder $\varphi_k$ first applies a window-level convolution to compress the input history $X_k\in\mathbb{R}^{T\times d}$ into a latent window representation.
This representation is then concatenated with temporal context embeddings (i.e., time-of-day and day-of-week) and processed by a stack of three-layer residual MLP blocks implemented with point-wise convolutions, yielding the latent representation $z_k\in\mathbb{R}^{D}$.
The predictor $\psi_k$ is a 1-layer $1{\times}1$ convolution that maps the invariant codebook representation
$\hat z_k$ to the $H$-step forecast $\hat Y_k\in \mathbb{R}^H$.

\noindent\textbf{Training and Inference.}
The training process of \method is elaborated in Algorithm~\ref{alg:fed}.
In each federated round, the server distributes the global codebook $Q$ to all clients (Line~3).
These clients initialize the local codebook $Q_k$ with $Q$ and update local encoder $\varphi_k$, codebook $Q_k$, and decoder $\psi_k$, with the supervised loss defined as:
\begin{equation}
  \mathcal{L}=\left\|\hat Y_k - Y_k\right\|_2^2 .
  \label{eq:local-obj}
\end{equation}
Then all clients upload the updated codebook $Q_k$ to the server (Lines~5-7), which perform alignment and aggregation to produce the novel global prototype $Q$ for the next round (Lines~8 and 9).

After training, each client is equipped with the global codebook $Q$ and personalized encoder and decoder. Therefore, they merely conduct the forward process to forecast the future horizons given the historical context, without any communication.

\begin{algorithm}[!h]
\caption{Training process of \method}
\label{alg:fed}
Server initializes $Q\in\mathbb{R}^{I\times D}$\;
\For{round $r=1,\dots,R$}{
  Server distributes $Q$ to all clients\;
  \For{client $k=1,\dots,N$ \emph{in parallel}}{
    $Q_k\gets Q$\;
    update $(\varphi_k,\psi_k,Q_k)$ in the loss function of Eq.~\eqref{eq:local-obj}
    , with $\hat Y_k= \psi_k(\hat z_k),\hat z_k=A(z_k;Q_k),z_k=\varphi_k(X_k)$;\\
    upload $Q_k$ \tcp*{$I\times D$ parameters}
  }
  Server: $\Pi_k\gets\operatorname{align}\big(Q_k,Q\big)$ for all $k$ following Eq.~\eqref{eq:assign-prob}\;
  Server: $Q\gets\textstyle\sum_k w_k\,\Pi_kQ_k$
}
\Return $Q$ ; $\{\varphi_k,\psi_k\}_{k=1}^N$
\end{algorithm}


\section{Experiments}
\label{sec:experiments}

We design experiments to answer the following research questions.
\begin{itemize}[leftmargin=*]
    \item \textbf{RQ1:} How does \method perform compared with existing federated and centralized STF baselines?
    \item \textbf{RQ2:} How different components contribute to \method?
    \item \textbf{RQ3:} Does the prototype codebook capture common environmental confounding effects transferable to unseen clients?
    \item \textbf{RQ4:} How efficient and robust is the prototype codebook?
    \item \textbf{RQ5:} Can the learned prototype codebook capture meaningful environmental regimes?
\end{itemize}

\begin{table*}[!t]
  \small
  \centering
  \caption{Comparison results of three groups of baselines on five benchmarks. 
  The best results in Group (i) are highlighted in \ybest{yellow}. The best and second-best results in Group (ii) and (iii) are marked in \best{underline} and \second{bold}, respectively.}
  \label{tab:main}
  \setlength{\tabcolsep}{0.055cm}
  \renewcommand\arraystretch{1.2}
  \begin{tabularx}{\textwidth}{@{}llcccccccccccccccc@{}}
    \toprule
    \multirow{2}{*}{Dataset} & \multirow{2}{*}{Metric} & \multicolumn{6}{c}{\textbf{Group~(i)}}
    & \multicolumn{3}{c}{\textbf{Group~(ii)}}
    & \multicolumn{7}{c}{\textbf{Group~(iii)}} \\
    \cmidrule(lr){3-8}\cmidrule(lr){9-11}\cmidrule(lr){12-18}
    & & DCRNN & GWNet & STID & CaST & STDN & FaST
    & FedAvg & FedProx & FedPer
    & FedGRU & FedGTP & CNFGNN & Fuels & FedDis & SC-FSGL & \textbf{\method} \\
    \midrule
    \multirow{2}{*}{{METR-LA}} & MAE  & \ybest{3.05} & 3.07 & 3.11 & 3.79 & 3.57 & 4.15 & 3.91 & 3.92 & 3.73 & 4.26 & 4.86 & 4.12 & \second{3.54} & 3.55 & 4.20 & \best{3.30} \\
                                     & RMSE & 6.30 & \ybest{6.27} & 6.49 & 7.44 & 7.31 & 9.57 & 7.53 & 7.55 & 7.41 & 8.12 & 10.51 & 7.89 & \second{6.88} & 7.01 & 9.62 & \best{6.93} \\
    \multirow{2}{*}{{PEMS-BAY}}& MAE  & 1.59 & 1.62 & \ybest{1.56} & 1.77 & 1.86 & 1.72 & 1.93 & 1.93 & 1.90 & 2.06 & 1.75 & 2.07 & 1.78 & 1.80 & \second{1.73} & \best{1.68} \\
                                     & RMSE & 3.69 & 3.71 & \ybest{3.59} & 3.97 & 4.43 & 4.18 & 4.34 & 4.36 & 4.32 & 4.79 & 3.98 & 4.79 & 4.01 & \second{3.94} & 3.98 & \best{3.91} \\
    \multirow{2}{*}{{PEMS03}}  & MAE  & 15.53 & \ybest{14.70} & 15.36 & 18.36 & 18.59& 15.42 & 17.73& 17.75& 17.75& 20.02& 17.31& 20.67& 16.23& \second{16.17} & 16.84& \best{15.79} \\
                                     & RMSE & 27.42 & 25.99 & 27.46 & 28.07 & 30.21& \ybest{25.08} & 27.66& 27.72& 27.80& 32.35& 27.98& 33.17& 26.51& \second{26.47} & 28.17& \best{26.32} \\
    \multirow{2}{*}{{PEMS04}}  & MAE  & 19.77 & 18.61 & \ybest{18.41} & 29.78 & 21.77& 19.85 & 23.28& 23.45& 22.82& 26.36& 22.15& 26.23& 20.12& 20.66& \second{19.67}& \best{19.14} \\
                                     & RMSE & 31.35 & 30.00 & \ybest{29.93} & 43.92 & 33.21& 32.52 & 36.27& 36.51& 35.73& 40.43& 34.17& 39.85& 32.07& \second{31.37} & 31.45& \best{30.82} \\
    \multirow{2}{*}{{KnowAir}} & MAE  & 15.54& \ybest{13.88}& 15.23& 14.86 & 14.01& 14.89 & 35.11& 36.94& 23.45& 26.64& 17.96& 26.67 & 17.82& 26.22& \second{17.28}& \best{15.07} \\
                                     & RMSE & 23.89& 22.11& 23.65& \ybest{16.38} & 22.33& 23.83 & 41.78& 43.74& 30.76& 34.13& 29.12& 34.15 & 26.27& 33.58 & \second{25.68}& \best{23.72} \\
    \bottomrule
  \end{tabularx}
\end{table*}

\subsection{Experimental Setup}
\label{sec:exp-setup}

\textbf{Datasets.}
We conduct experiments on five widely-used benchmarking datasets:
METR-LA~\citep{li2018diffusion}, PEMS-BAY~\citep{li2018diffusion}, PEMS03~\citep{song2020spatial}, PEMS04~\citep{song2020spatial}, and KnowAir~\citep{wang2020pm2}.
METR-LA and PEMS-BAY include traffic speed recordings with the time interval of 5 minutes collected from 207 loop detectors 
from March, 2012 to June, 2012 in Los Angeles and 325 loop detectors from January, 2017 to May, 2017
in the Bay Area, respectively.
PEMS03 and PEMS04 collect traffic flow recordings with the time interval of 5 minutes from 358 sensors from September, 2018 to November, 2018 and 307 sensors from January, 2018 to February, 2018 in California, respectively.
KnowAir includes the PM2.5 recordings of 184 cities in China.

\noindent\textbf{Baselines.}
We compare \method with the following three groups of baselines:
\textbf{(i) Centralized Methods}: 
DCRNN~\citep{li2018diffusion}, GWNet~\citep{wu2019graphwavenet},
STID~\citep{shao2022stid}, CaST~\citep{xia2023cast}, STDN~\citep{cao2025spatiotemporal}, and
FaST~\citep{zhao2026fast};
\textbf{(ii) General Federated Methods}:
FedAvg~\citep{mcmahan2017fedavg},
FedProx~\citep{li2020fedprox},
and FedPer~\citep{arivazhagan2019fedper};
\textbf{(iii) Federated Spatio-Temporal Methods}:
FedGRU~\citep{liu2020fedgru},
FedGTP~\citep{yang2024fedgtp},
CNFGNN~\citep{meng2021cnfgnn},
Fuels~\citep{fuels2025},
FedDis~\citep{feddis}, and
SC-FSGL~\citep{SCFSGL}.
For the general federated methods, we adopt the same encoder and decoder architecture with \method, and refer to the original code repository for centralized and federated spatio-temporal methods.

\noindent\textbf{Implementation Details.}
We term each node as a client in \method. 
The length of historical and prediction windows is set to~12.
Data are split chronologically into training, validation, and test datasets as
$7{:}1{:}2$ (for {METR-LA}, {PEMS-BAY}, and KnowAir) and $6{:}2{:}2$
(for {PEMS03} and {PEMS04})~\citep{shao2024exploring}.
The \method uses $I{=}10$ prototypes and a 3-layer ResMLP in encoder, with $D{=}64$.
The temperature $\alpha$ is set as 0.1.
All clients are selected in each round and perform~1 epoch local update using Adam optimizer with the initial learning rate of 0.001 and a batch size of 256.
The federated round $R$ is set to 200.
The effects of different hyperparameter settings on the performance is investigated in Appendix~\ref{sec:app-param}. All experiments are implemented with PyTorch 2.8 on a server with NVIDIA A6000.

\subsection{RQ1: Overall Forecasting Performance}
\label{sec:exp-main}
We report the numerical results of different baselines in Table~\ref{tab:main}.
The best-performed baselines in Groups~(ii) and~(iii) vary across different datasets, while \method consistently outperforms all federated baselines, demonstrating the effectiveness of the proposed federated de-confounding mechanism.
The methods specialized for spatio-temporal heterogeneity (i.e., \method, FedDis, and SC-FSGL) consistently achieve more performance gains than general federated learning methods. 
These general federated learning methods (e.g., FedAvg, FedProx, and FedPer) exhibit substantially higher errors despite adopting the same encoder-decoder architectures with~\method.
These methods learn a global representation space through parameter averaging, implicitly pushing heterogeneous clients toward a consensus parameterization that may smooth out client-specific temporal dynamics.
Our proposed \method shares only the prototype codebook that captures latent environmental regimes across clients, while allowing each client to retain its own encoder and prediction head.
This design enables the shared prototype codebook to exploit cross-client environmental heterogeneity without sacrificing client-specific forecasting patterns.
Compared with centralized methods, which have access to the collective training data, \method narrows the performance gap to within 4.9\%--6.7\% on METR-LA and 3.2\%--3.3\% on PEMS-BAY.
These results suggest that sharing latent environmental regimes, rather than averaging model parameters or exchanging raw observations, is sufficient to approach centralized performance in federated spatio-temporal forecasting.

\begin{table}[t]
  \centering
  \caption{Ablation study on METR-LA and PEMS-BAY. The best result in each row is highlighted in \best{underline}.}
  \label{tab:ablation}
  \setlength{\tabcolsep}{0.13cm}
  \renewcommand\arraystretch{1}
  \begin{tabularx}{\linewidth}{@{}llccccccc@{}}
    \toprule
    Dataset & Metric & \method & A.1 & A.2 & A.3 & A.4 & A.5 & A.6 \\
    \midrule
    \multirow{2}{*}{METR-LA} & MAE  & \best{3.30} & 5.85 & 4.09 & 11.92 & 4.01 & 5.09 & 3.69 \\
                              & RMSE & \best{6.93} & 8.92 & 7.81 & 13.26 & 8.22 & 9.54 & 7.21 \\
    \multirow{2}{*}{PEMS-BAY} & MAE  & \best{1.68} & 4.00 & 3.01 & 8.81 & 2.04 & 3.06 & 1.91 \\
                              & RMSE & \best{3.91} & 6.71 & 6.52 & 10.16 & 4.80 & 6.65 & 4.08 \\
    \bottomrule
  \end{tabularx}
\end{table}

\subsection{RQ2: Contribution Analysis in \method}
To assess the contribution of each component, we compare six ablated variants against the full \method in Table~\ref{tab:ablation}.
In \textbf{A.1}, we remove the federated prototype aggregation and train each client independently using only local data.
Compared with \method, MAE increases by $77\%$ on METR-LA and $138\%$ on PEMS-BAY.
This substantial degradation indicates that a single client cannot adequately capture the stable environmental regimes. Federated prototype aggregation makes clients benefit from shared latent environmental knowledge learned across heterogeneous environments.
In \textbf{A.2} and \textbf{A.3}, we further share the encoder and the entire encoder--decoder architecture across clients, respectively.
Both variants lead to remarkable performance degradation, demonstrating that forcing heterogeneous clients into a unified representation space collapses client-specific temporal modeling, even though the global environmental regimes can still be accessible.
In \textbf{A.4}, we remove the prototype alignment procedure before aggregation, resulting in a consistent performance drop.
This result verifies that prototype alignment is necessary for preserving the consistency of environmental semantics across clients before aggregation.
In \textbf{A.5}, we replace the soft prototype assignment in Eq.~(\ref{eq:assign}) with hard assignment.
The degradation suggests that soft assignment is essential for smoothly modeling transitions between latent environmental regimes. The results are consistent with the intuition that temporal observations are jointly affected by various environmental regimes.
Finally, in \textbf{A.6}, each client uploads only its dominant prototype (i.e., the prototype with the largest assignment weight) instead of the complete prototype update.
The inferior performance indicates that the global environmental regimes are encoded by multiple prototypes, and cannot be adequately represented through sparse prototype communication.

\begin{table}[!t]
  \centering
  \caption{The RMSE results of held-out clients under varying sampling fractions. The best result in each column is highlighted in \best{underline}.}
  \label{tab:transfer}
  \setlength{\tabcolsep}{0.22cm}
  \renewcommand\arraystretch{1.05}
  \begin{tabularx}{\linewidth}{@{}llccccc@{}}
    \toprule
    Dataset & Method & 5\% & 25\% & 50\% & 75\% & 90\% \\
    \midrule
    \multirow{4}{*}{METR-LA}
    & Frozen $Q$ & 6.98 & 7.19 & 7.08 & 7.46 & 7.68 \\
    & Random $Q_k$ & 9.10 & 9.14 & 9.14 & 9.22 & 9.20 \\
    & $\method_{FT}$  & 6.98 & 7.15 & 7.05 & 7.38 & 7.68 \\
    & $\method_{Bar}$       & \best{6.40} & \best{6.56} & \best{6.50} & \best{6.70} & \best{7.10} \\
    \midrule
    \multirow{4}{*}{PEMS-BAY}
    & Frozen $Q$ & 3.90 & 3.86 & 3.87 & 3.96 & 3.86 \\
    & Random $Q_k$ & 6.69 & 6.73 & 6.53 & 6.58 & 6.54 \\
    & $\method_{FT}$ & 3.89 & 3.86 & 3.88 & 3.97 & 3.81 \\
    & $\method_{Bar}$       & \best{3.62} & \best{3.58} & \best{3.54} & \best{3.53} & \best{3.61} \\
    \bottomrule
  \end{tabularx}
\end{table}

\subsection{RQ3: Transferability to Unseen Clients}

We investigate whether the shared prototype codebook captures transferable latent environmental regimes rather than memorizing those of the participation clients.
For each participation rate $g\in\{5\%,25\%,50\%,75\%,90\%\}$, we first pretrain the prototype codebook using only the selected $g\%$ clients.
As shown in Table~\ref{tab:transfer}, the remaining $(1-g)\%$ held-out clients are then evaluated under four settings:
\emph{(i)} {Frozen $Q$}: the pretrained codebook is frozen and transferred to the held-out clients for local training;
\emph{(ii)} {Random $Q_k$}: each held-out client initializes its codebook randomly and trains locally without federation;
\emph{(iii)} $\method_{FT}$: the held-out clients collaboratively learn a new codebook from scratch through federated training; and
\emph{(iv)} $\method_{Bar}$: the performance of the held-out clients when all clients participate in federated training.

Table~\ref{tab:transfer} shows that the codebook learned from only a subset of clients transfers effectively to unseen clients.
The advantage is particularly evident when $s=5\%$, where the codebook is pretrained using only 10 METR-LA clients or 16 PEMS-BAY clients, yet Frozen $Q$ achieves performance comparable to $\method_{FT}$.
In contrast, Random $Q_k$ performs substantially worse than Frozen $Q$ on both datasets.
Therefore, the transferred codebook provides reusable latent environmental prototypes that unseen clients can exploit through lightweight local adaptation.
The remaining gap between Frozen $Q$ and $\method_{Bar}$ is expected, since the latter learns the shared codebook from all participating clients and hence captures a broader range of environmental conditions.
Overall, these results suggest that the learned prototype codebook captures transferable latent environmental regimes, which can be effectively reused by distinct nodes.

\begin{figure}[!t]
  \centering
  \includegraphics[width=\linewidth]{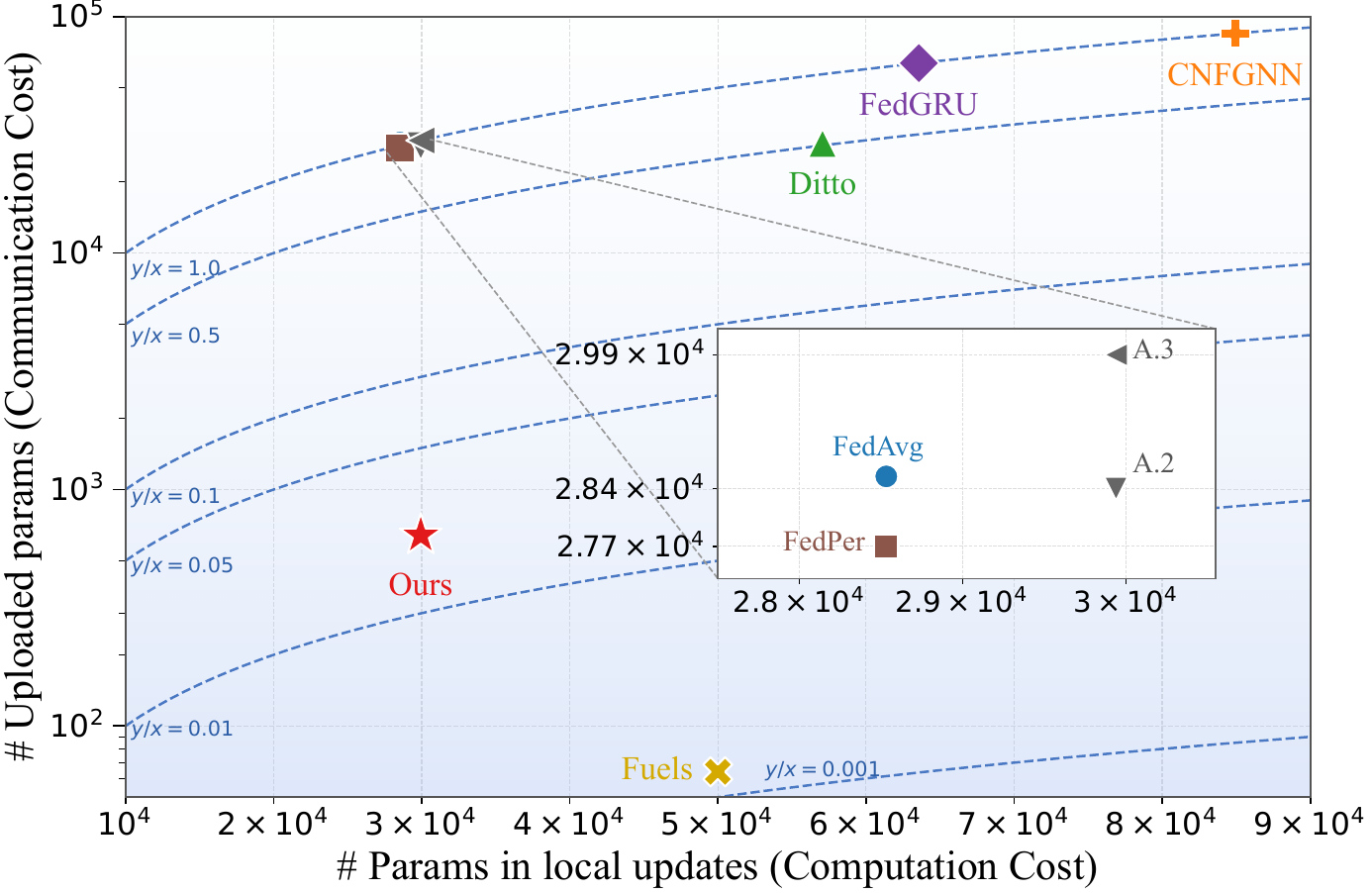}
  \caption{Trade-off between local updates and uploads. Dashed curves show different upload-to-update ratios.}
  \label{fig:comm-efficiency}
\end{figure}

\subsection{RQ4: Efficiency and Robustness}
\subsubsection{Efficiency Analysis}

We compare the computational and communication costs of different federated methods in Figure~\ref{fig:comm-efficiency}.
The horizontal axis denotes the number of locally updated parameters and the vertical axis denotes the number of uploaded parameters per communication round in each client.
In \method, each client uploads only the $I\times D$ prototype codebook ($640$ floating-point values with $I{=}10$ and $D{=}64$), regardless of the encoder backbone size.
This communication cost is approximately two orders of magnitude smaller than that of FedAvg, FedGRU, and CNFGNN, all of which upload the complete model parameters.
FedPer slightly reduces the upload cost by keeping the prediction head local.
Fuels achieves the smallest communication overhead (64 values) through a compact latent representation, at the expense of substantially larger local parameter updates.
In A.2, local decoders are preserved, and therefore it generates moderate lower communication cost compared with FedAvg. 
However, A.3 has to exchange additional prototype codebook, thereby leading to higher communication cost.
Overall, \method achieves a favorable trade-off between local computation and communication efficiency, requiring only lightweight codebook exchange while consistently delivering the strongest forecasting performance among federated baselines.

\begin{table}[!t]
  \centering
  \caption{Robustness of prototypes under different noise types. The best result in each row is highlighted in \best{underline}.}
  \label{tab:codebook-noise}
  \setlength{\tabcolsep}{0.1cm}
  \renewcommand\arraystretch{1}
  \begin{tabularx}{\linewidth}{@{}llccccc@{}}
    \toprule
    Dataset & Metric & \method & +Gaussian & +Laplace & +Rademacher \\
    \midrule
    \multirow{2}{*}{METR-LA} & MAE  & \best{3.30} & 3.39 & 3.48 & 3.36 \\
                              & RMSE & \best{6.93} & 7.08 & 7.21 & 7.02 \\
    \multirow{2}{*}{PEMS-BAY} & MAE  & \best{1.68} & 1.69 & 2.41 & 1.69 \\
                              & RMSE & \best{3.91} & 3.92 & 5.74 & 3.91 \\
    \bottomrule
  \end{tabularx}
\end{table}

\subsubsection{Robustness of Codebooks}
To evaluate the robustness of the learned prototype codebook, we inject Gaussian noise ($\mu=0,\sigma=0.05$), Laplace noise ($b=0.05$), and Rademacher noise ($\pm0.05$) into the learned codebook after local training.
The numerical results are reported in Table~\ref{tab:codebook-noise}.
The learned codebook exhibits strong robustness to Gaussian and Rademacher perturbations, with MAE increasing by less than $0.09$ on METR-LA and less than $0.01$ on PEMS-BAY.
In comparison, Laplace perturbation leads to a little more degradation, with the MAE on PEMS-BAY increasing by $0.73$.
The results indicates that the learned prototype codebook captures stable environmental regimes and the injected noise does not fundamentally perturb the environmental semantics encoded by the prototypes.
Therefore, the proposed \method promises the incorporation of diverse differential privacy approaches to further avoid the information leakage from the uploaded codebooks.

\begin{figure}[!t]
  \includegraphics[width=\linewidth]{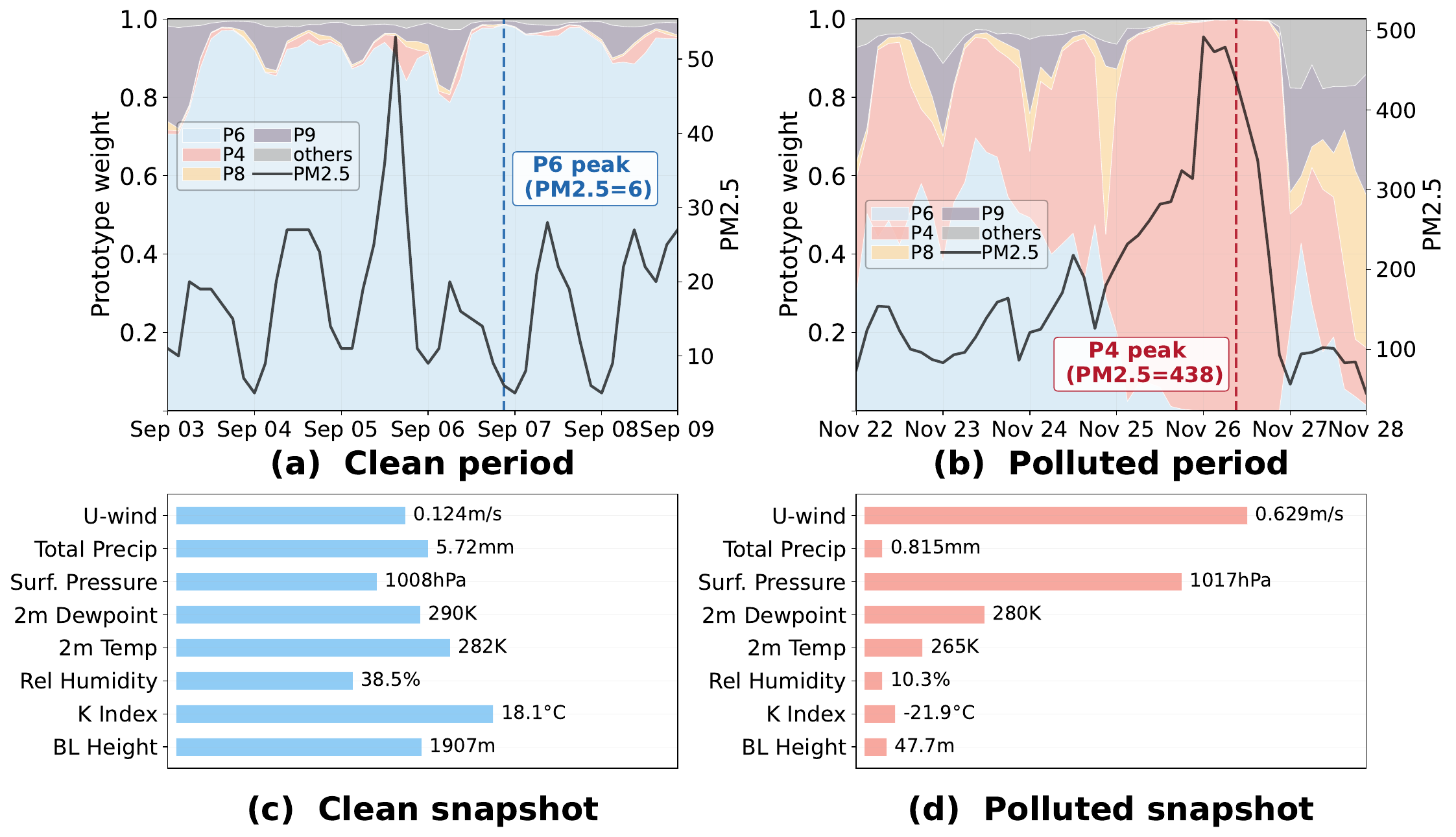}
  \caption{Prototype interpretation on the KnowAir dataset.
(a)--(b) Prototype assignment weights and PM$_{2.5}$ distribution during clean and polluted periods at Station~6.
(c)--(d) Corresponding meteorological observations.
The prototype activations change consistently with the underlying environmental conditions, indicating that the learned codebook captures latent environmental regimes rather than only the target variable.}
\label{fig:codebook-interpretation}
\end{figure}

\subsection{RQ5: Understanding the Codebook}
\label{sec:case-study}
\subsubsection{Interpreting the codebook}
In this subsection, we verify that the learned prototype codebook captures interpretable environmental regimes rather than merely tracking the target variable.
We present two contrastive 48-step input windows from Station~6 on KnowAir dataset, including a clean period (mean PM$_{2.5}=17.6\,\mu$g/m$^3$) and a heavily polluted period (mean PM$_{2.5}=184.5\,\mu$g/m$^3$).
Figure~\ref{fig:codebook-interpretation} presents the prototype assignment weights and corresponding meteorological conditions.

As shown in Figure~\ref{fig:codebook-interpretation}(a) and (b), the prototype assignment patterns differ substantially between the two periods.
Prototype~\texttt{P6} dominates the clean window but decreases to~0.28 in the polluted window, whereas~\texttt{P4} increases from~0.01 to~0.53.
The $L_1$ distance between the two assignment vectors is 1.26 and their cosine similarity is 0.48, indicating distinct latent environmental regimes.
The meteorological snapshots in Figure~\ref{fig:codebook-interpretation}(c) and (d) further show that the difference is not solely caused by PM$_{2.5}$.
The temperature, surface pressure, wind, and humidity exhibit consistent change between the two periods, suggesting that the learned prototypes capture broader environmental contexts rather than only target-variable patterns.

\subsubsection{Interpreting Prototype Assignments}

\begin{figure}[!t]
  \includegraphics[width=\linewidth]{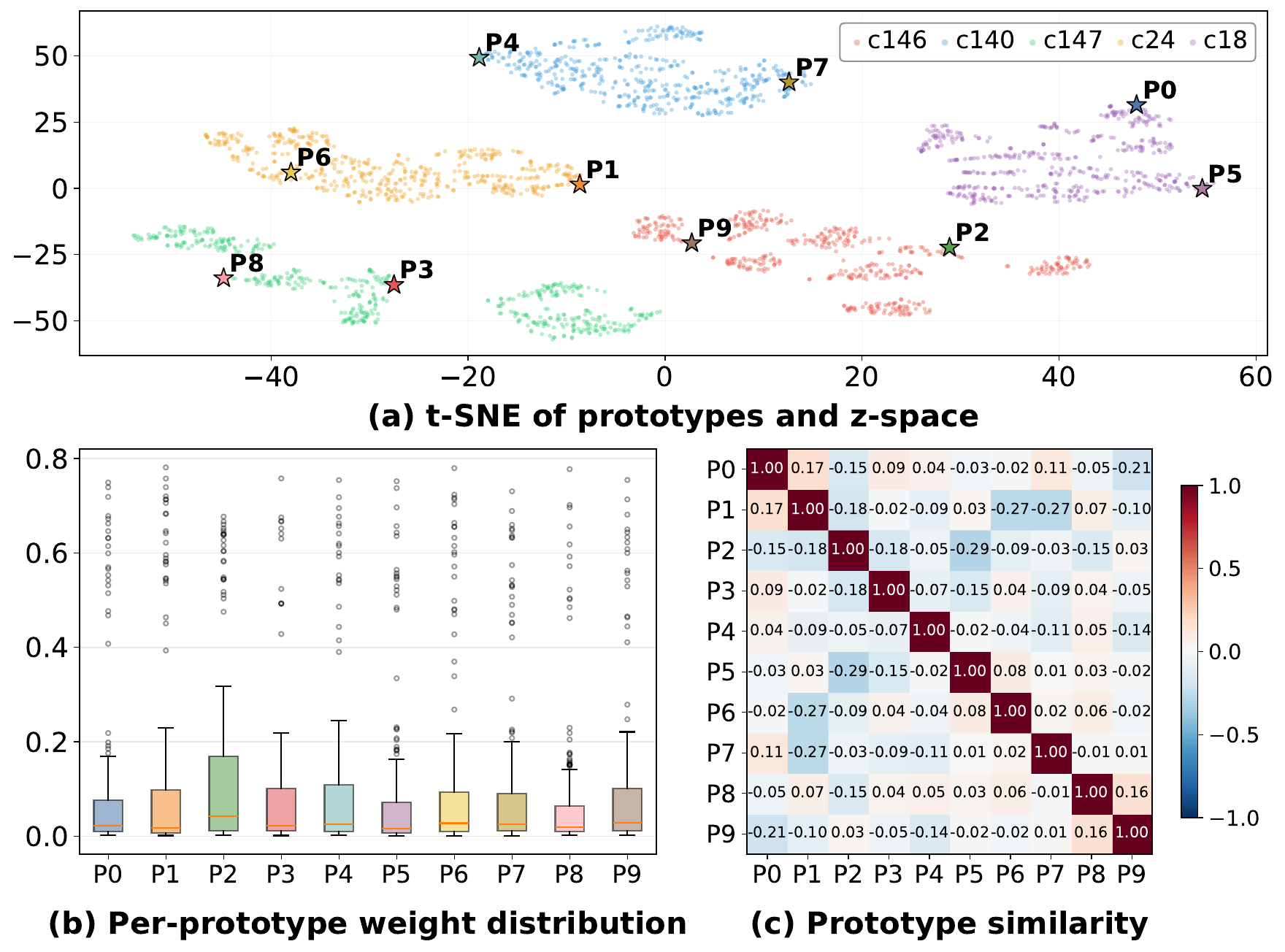}
  \caption{Visualization of learned representations and prototype assignments on the KnowAir dataset.
(a) $t$-SNE visualization of encoder outputs and global prototype vectors.
(b) Distribution of prototype assignment weights across 184 clients.
(c) Pairwise cosine similarity matrix among prototypes.
The results show that prototypes capture diverse latent environmental patterns without collapsing into redundant representations.}
  \label{fig:tsne-case-study}
\end{figure}

In Figure~\ref{fig:tsne-case-study}(a), we visualize the $t$-SNE of prototypes and encoder outputs from five stations.
The learned representations preserve heterogeneous environmental patterns rather than collapsing into a shared representation space.
The global prototype vectors are distributed across the representation manifold and located near different groups of local representations.
It suggests that \method cultivates shared environmental regimes, enabling clients to align through soft weights without directly aggregating local encoder parameters.
Figure~\ref{fig:tsne-case-study}(b) further shows that soft-assignment weights vary substantially across the 184 clients.
Specifically, prototypes \texttt{P1} and \texttt{P2} exhibit the largest cross-client variance, suggesting that different prototypes capture environmental patterns that are more prevalent in certain regions.
The average maximum assignment weight across all clients is 0.59, indicating that each client typically relies on a small subset of prototypes rather than uniformly activating the entire codebook.
This observation supports the interpretation that the prototype codebook provides a compact representation of diverse environmental regimes.
Finally, the pairwise cosine similarity matrix in Figure~\ref{fig:tsne-case-study}(c) shows low similarity among different prototype vectors.
This indicates that the learned codebook maintains sufficient diversity and avoids redundant prototypes, allowing the $I=10$ prototypes to represent different latent environmental patterns.


\section{Conclusion and Discussion}
In this paper, we revisit the spatio-temporal heterogeneity from the perspective of causal de-confounding. 
We propose \method as a novel federated de-confounding framework to treat clients as distinct causal environments and learn a global prototype codebook for the shared environmental regimes.
We theoretically analyze that the federated de-confounding bound is linearly dependent on the averaged confounding strengths.
We conduct extensive empirical experiments to validate the state-of-the-art performance of \method, with great transferability, robustness, interpretability and communication efficiency.

\noindent\textbf{Limitations and future works:}:
The discrete prototypes may struggle with continuous or extremely high-dimensional shifts. Additionally, the $O(I^3)$ alignment complexity may become a bottleneck if the prototype scale increases significantly.
Future research will explore scalable alignment strategies and dynamic codebooks. We also aim to expand \method toward building robust urban foundation models across heterogeneous sensor networks.

\section*{Ethical Statement}
This research contributes to the advancement of privacy-preserving spatio-temporal forecasting. By utilizing a federated learning paradigm, \method enables collaborative urban modeling without centralizing sensitive raw data, thereby mitigating risks of data leakage and aligning with global privacy regulations. 
All experiments were conducted using publicly available benchmark datasets.
We acknowledge the use of Large Language Models during the preparation of this manuscript for the purpose of language polishing, grammar correction, and structural refinement. All core scientific contributions are developed solely by the authors.
\bibliographystyle{ACM-Reference-Format}
\bibliography{software}

@inproceedings{shao2022stid,
  title={Spatial-Temporal Identity: A Simple yet Effective Baseline for Multivariate Time Series Forecasting},
  author={Shao, Zezhi and Zhang, Zhao and Wang, Fei and Wei, Wei and Xu, Yongjun},
  booktitle={CIKM}, year={2022}}

@article{kuhn1955hungarian,
  title={The {Hungarian} method for the assignment problem},
  author={Kuhn, Harold W.},
  journal={Naval Research Logistics Quarterly},
  volume={2}, number={1-2}, pages={83--97}, year={1955}}

@article{arjovsky2019irm,
  title={Invariant Risk Minimization},
  author={Arjovsky, Martin and Bottou, L{\'e}on and Gulrajani, Ishaan and Lopez-Paz, David},
  journal={arXiv:1907.02893}, year={2019}}

@article{peters2016icp,
  title={Causal Inference by using Invariant Prediction: Identification and Confidence Intervals},
  author={Peters, Jonas and B{\"u}hlmann, Peter and Meinshausen, Nicolai},
  journal={J. R. Stat. Soc. B}, volume={78}, number={5}, pages={947--1012}, year={2016}}

@book{pearl2009causality,
  title={Causality},
  author={Pearl, Judea},
  year={2009},
  publisher={Cambridge university press}
}

@inproceedings{li2017dcrnn,
  title={Diffusion Convolutional Recurrent Neural Network: Data-Driven Traffic Forecasting},
  author={Li, Yaguang and Yu, Rose and Shahabi, Cyrus and Liu, Yan},
  booktitle={ICLR}, year={2018}}

@inproceedings{wu2019graphwavenet,
  title={Graph WaveNet for Deep Spatial-Temporal Graph Modeling},
  author={Wu, Zonghan and Pan, Shirui and Long, Guodong and Jiang, Jing and Zhang, Chengqi},
  booktitle={IJCAI}, year={2019}}

@article{liu2020fedgru,
  title={Privacy-preserving Traffic Flow Prediction: A Federated Learning Approach},
  author={Liu, Yi and Yu, James J. Q. and Kang, Jiawen and Niyato, Dusit and Zhang, Shuyu},
  journal={IEEE Internet of Things Journal}, volume={7}, number={8}, pages={7751--7763}, year={2020}}

@inproceedings{meng2021cnfgnn,
  title={Cross-Node Federated Graph Neural Network for Spatio-Temporal Data Modeling},
  author={Meng, Chuizheng and Rambhatla, Sirisha and Liu, Yan},
  booktitle={KDD}, year={2021}}

@inproceedings{fuels2025,
  title={Personalized Federated Learning for Spatio-Temporal Forecasting: A Dual Semantic Alignment-Based Contrastive Approach},
  author={Liu, Qingxiang and Sun, Sheng and Liang, Yuxuan and Xue, Jingjing and Liu, Min},
  booktitle={AAAI}, pages={12192--12200}, year={2025}}

@inproceedings{feddis,
  title={Feddis: A causal disentanglement framework for federated traffic prediction},
  author={Zhou, Chengyang and Zhang, Zijian and Zhang, Chunxu and Miao, Hao and Zhang, Yulin and Lyu, Kedi and Hu, Juncheng},
  booktitle={Proceedings of the ACM Web Conference 2026},
  pages={7541--7551},
  year={2026}
}

@inproceedings{yang2024fedgtp,
  title={FedGTP: exploiting inter-client spatial dependency in federated graph-based traffic prediction},
  author={Yang, Linghua and Chen, Wantong and He, Xiaoxi and Wei, Shuyue and Xu, Yi and Zhou, Zimu and Tong, Yongxin},
  booktitle={Proceedings of the 30th ACM SIGKDD conference on knowledge discovery and data mining},
  pages={6105--6116},
  year={2024}
}

@inproceedings{xia2023cast,
  title={Deciphering Spatio-Temporal Graph Forecasting: A Causal Lens and Treatment},
  author={Xia, Yutong and Liang, Yuxuan and Wen, Haomin and Liu, Xu and Wang, Kun and Zhou, Zhengyang and Zimmermann, Roger},
  booktitle={NeurIPS}, year={2023}}

@article{capaint2024,
  title={Causal Deciphering and Inpainting in Spatio-Temporal Dynamics via Diffusion Model},
  author={Duan, Yifan and Zhao, Jian and Mao, Junyuan and Wu, Hao and Xu, Jingyu and Wang, Shilong and Ma, Caoyuan and Wang, Kai and Wang, Kun and Li, Xuelong},
  journal={arXiv preprint arXiv:2409.19608}, year={2024}}

@inproceedings{zhou2023caustg,
  title={Maintaining the Status Quo: Capturing Invariant Relations for OOD Spatiotemporal Learning},
  author={Zhou, Zhengyang and Huang, Qihe and Yang, Kuo and Wang, Kun and Wang, Xu and Zhang, Yudong and Liang, Yuxuan and Wang, Yang},
  booktitle={KDD}, pages={3603--3614}, year={2023}}

@inproceedings{wang2024stone,
  title={{STONE}: A Spatio-temporal OOD Learning Framework Kills Both Spatial and Temporal Shifts},
  author={Wang, Binwu and Ma, Jiaming and Wang, Pengkun and Wang, Xu and Zhang, Yudong and Zhou, Zhengyang and Wang, Yang},
  booktitle={KDD}, year={2024}}

@inproceedings{ji2025seeing,
  title={Seeing the unseen: Learning basis confounder representations for robust traffic prediction},
  author={Ji, Jiahao and Zhang, Wentao and Wang, Jingyuan and Huang, Chao},
  booktitle={Proceedings of the 31st ACM SIGKDD Conference on Knowledge Discovery and Data Mining V. 1},
  pages={577--588},
  year={2025}
}

@inproceedings{yuan2023eagle,
  title={Environment-Aware Dynamic Graph Learning for Out-of-Distribution Generalization},
  author={Yuan, Haonan and Sun, Qingyun and Fu, Xingcheng and Zhang, Ziwei and Ji, Cheng and Peng, Hao and Li, Jianxin},
  booktitle={NeurIPS}, year={2023}}

@inproceedings{wang2024nuwadynamics,
  title={{NuwaDynamics}: Discovering and Updating in Causal Spatio-Temporal Modeling},
  author={Wang, Kun and Wu, Hao and Duan, Yifan and Zhang, Guibin and Wang, Kai and Peng, Xiaojiang and Zheng, Yu and Liang, Yuxuan and Wang, Yang},
  booktitle={ICLR}, year={2024}}

@inproceedings{ma2025robust,
  title={Robust spatio-temporal centralized interaction for ood learning},
  author={Ma, Jiaming and Wang, Binwu and Wang, Pengkun and Zhou, Zhengyang and Wang, Xu and Wang, Yang},
  booktitle={Forty-second International Conference on Machine Learning},
  year={2025}
}

@inproceedings{yang2025revealing,
  title={Revealing concept shift in spatio-temporal graphs via state learning},
  author={Yang, Kuo and Guo, Yunhe and Huang, Qihe and Zhou, Zhengyang and Wang, Yang},
  booktitle={Proceedings of the Thirty-Fourth International Joint Conference on Artificial Intelligence},
  pages={3525--3533},
  year={2025}
}

@inproceedings{SCFSGL,
  title={Causality-inspired federated learning for dynamic spatio-temporal graphs},
  author={Liu, Yuxuan and Xu, Wenchao and Wang, Haozhao and He, Zhiming and Shi, Zhaofeng and Xu, Chongyang and Wang, Peichao and Zhang, Boyuan},
  booktitle={Proceedings of the AAAI Conference on Artificial Intelligence},
  volume={40},
  number={28},
  pages={23926--23934},
  year={2026}
}

@article{qi2025federated,
  title={Federated deconfounding and debiasing learning for out-of-distribution generalization},
  author={Qi, Zhuang and Zhou, Sijin and Meng, Lei and Hu, Han and Yu, Han and Meng, Xiangxu},
  journal={arXiv preprint arXiv:2505.04979},
  year={2025}
}

@inproceedings{tang2024learning,
  title={Learning personalized causally invariant representations for heterogeneous federated clients},
  author={Tang, Xueyang and Guo, Song and Zhang, Jie and Guo, Jingcai},
  booktitle={International Conference on Learning Representations},
  volume={2024},
  pages={10016--10037},
  year={2024}
}

@inproceedings{tang2024causally,
  title={Causally motivated personalized federated invariant learning with shortcut-averse information-theoretic regularization},
  author={Tang, Xueyang and Guo, Song and Guo, Jingcai and Zhang, Jie and Yu, Yue},
  booktitle={Forty-first International Conference on Machine Learning},
  year={2024}
}

@inproceedings{guo2025federated,
  title={Federated causally invariant feature learning},
  author={Guo, Xianjie and Yu, Kui and Cui, Lizhen and Yu, Han and Li, Xiaoxiao},
  booktitle={Proceedings of the AAAI Conference on Artificial Intelligence},
  volume={39},
  number={16},
  pages={16978--16986},
  year={2025}
}

@article{li2025federated,
  title={Federated domain generalization: A survey},
  author={Li, Ying and Wang, Xingwei and Zeng, Rongfei and Donta, Praveen Kumar and Murturi, Ilir and Huang, Min and Dustdar, Schahram},
  journal={Proceedings of the IEEE},
  year={2025},
  publisher={IEEE}
}

@inproceedings{mcmahan2017fedavg,
  title={Communication-Efficient Learning of Deep Networks from Decentralized Data},
  author={McMahan, Brendan and Moore, Eider and Ramage, Daniel and Hampson, Seth and Ag{\"u}era y Arcas, Blaise},
  booktitle={AISTATS}, year={2017}}

@inproceedings{li2020fedprox,
  title={Federated Optimization in Heterogeneous Networks},
  author={Li, Tian and Sahu, Anit Kumar and Zaheer, Manzil and Sanjabi, Maziar and Talwalkar, Ameet and Smith, Virginia},
  booktitle={MLSys}, year={2020}}

@article{arivazhagan2019fedper,
  title={Federated Learning with Personalization Layers},
  author={Arivazhagan, Manoj Ghuhan and Aggarwal, Vinay and Singh, Aaditya Kumar and Choudhary, Sunav},
  journal={arXiv preprint arXiv:1912.00818}, year={2019}}

@inproceedings{wang2020pm2,
  title={Pm2. 5-gnn: A domain knowledge enhanced graph neural network for pm2. 5 forecasting},
  author={Wang, Shuo and Li, Yanran and Zhang, Jiang and Meng, Qingye and Meng, Lingwei and Gao, Fei},
  booktitle={Proceedings of the 28th international conference on advances in geographic information systems},
  pages={163--166},
  year={2020}
}

@inproceedings{li2018diffusion,
  title={Diffusion Convolutional Recurrent Neural Network: Data-Driven Traffic Forecasting},
  author={Li, Yaguang and Yu, Rose and Shahabi, Cyrus and Liu, Yan},
  booktitle={ICLR},
  year={2018}
}

@inproceedings{song2020spatial,
  title={Spatial-temporal synchronous graph convolutional networks: A new framework for spatial-temporal network data forecasting},
  author={Song, Chao and Lin, Youfang and Guo, Shengnan and Wan, Huaiyu},
  booktitle={Proceedings of the AAAI conference on artificial intelligence},
  volume={34},
  number={01},
  pages={914--921},
  year={2020}
}

@inproceedings{cao2025spatiotemporal,
  title={Spatiotemporal-aware trend-seasonality decomposition network for traffic flow forecasting},
  author={Cao, Lingxiao and Wang, Bin and Jiang, Guiyuan and Yu, Yanwei and Dong, Junyu},
  booktitle={Proceedings of the AAAI Conference on Artificial Intelligence},
  volume={39},
  number={11},
  pages={11463--11471},
  year={2025}
}

@inproceedings{zhao2026fast,
  title={FaST: Efficient and Effective Long-Horizon Forecasting for Large-Scale Spatial-Temporal Graphs via Mixture-of-Experts},
  author={Zhao, Yiji and Zhong, Zihao and Wang, Ao and Wen, Haomin and Jin, Ming and Liang, Yuxuan and Wan, Huaiyu and Wu, Hao},
  booktitle={Proceedings of the 32nd ACM SIGKDD Conference on Knowledge Discovery and Data Mining V. 1},
  pages={1975--1986},
  year={2026}
}

@article{shao2024exploring,
 title={Exploring progress in multivariate time series forecasting: Comprehensive benchmarking and heterogeneity analysis},
 author={Shao, Zezhi and Wang, Fei and Xu, Yongjun and Wei, Wei and Yu, Chengqing and Zhang, Zhao and Yao, Di and Sun, Tao and Jin, Guangyin and Cao, Xin and others},
 journal={IEEE Transactions on Knowledge and Data Engineering},
 year={2024},
 volume={37},
 number={1},
 pages={291-305},
 publisher={IEEE}
}

@article{fang2026unraveling,
  title={Unraveling spatio-temporal foundation models via the pipeline lens: A comprehensive review},
  author={Fang, Yuchen and Miao, Hao and Liang, Yuxuan and Deng, Liwei and Cui, Yue and Zeng, Ximu and Xia, Yuyang and Zhao, Yan and Pedersen, Torben Bach and Jensen, Christian S and others},
  journal={IEEE Transactions on Knowledge and Data Engineering},
  year={2026},
  publisher={IEEE}
}

@article{mao2025survey,
  title={A survey on spatio-temporal prediction: From transformers to foundation models},
  author={Mao, Yingchi and Zhou, Hongliang and Chen, Ling and Qi, Rongzhi and Sun, Zhende and Rong, Yi and He, Xiaoming and Chen, Mingkai and Mumtaz, Shahid and Frascolla, Valerio and others},
  journal={ACM Computing Surveys},
  volume={58},
  number={4},
  pages={1--36},
  year={2025},
  publisher={ACM New York, NY}
}

@inproceedings{zhang2024urban,
  title={Urban foundation models: A survey},
  author={Zhang, Weijia and Han, Jindong and Xu, Zhao and Ni, Hang and Liu, Hao and Xiong, Hui},
  booktitle={Proceedings of the 30th ACM SIGKDD Conference on Knowledge Discovery and Data Mining},
  pages={6633--6643},
  year={2024}
}

@article{zhang2024modeling,
  title={Modeling spatio-temporal mobility across data silos via personalized federated learning},
  author={Zhang, Yudong and Wang, Xu and Wang, Pengkun and Wang, Binwu and Zhou, Zhengyang and Wang, Yang},
  journal={IEEE Transactions on Mobile Computing},
  volume={23},
  number={12},
  pages={15289--15306},
  year={2024},
  publisher={IEEE}
}

@article{graves2013generating,
  title={Generating sequences with recurrent neural networks},
  author={Graves, Alex},
  journal={arXiv preprint arXiv:1308.0850},
  year={2013}
}

@article{bai2018empirical,
  title={An empirical evaluation of generic convolutional and recurrent networks for sequence modeling},
  author={Bai, Shaojie and Kolter, J Zico and Koltun, Vladlen},
  journal={arXiv preprint arXiv:1803.01271},
  year={2018}
}

@article{kipf2016semi,
  title={Semi-supervised classification with graph convolutional networks},
  author={Kipf, Thomas N and Welling, Max},
  journal={arXiv preprint arXiv:1609.02907},
  year={2016}
}

@inproceedings{guo2019attention,
  title={Attention based spatial-temporal graph convolutional networks for traffic flow forecasting},
  author={Guo, Shengnan and Lin, Youfang and Feng, Ning and Song, Chao and Wan, Huaiyu},
  booktitle={Proceedings of the AAAI conference on artificial intelligence},
  volume={33},
  number={01},
  pages={922--929},
  year={2019}
}

@inproceedings{zheng2020gman,
  title={Gman: A graph multi-attention network for traffic prediction},
  author={Zheng, Chuanpan and Fan, Xiaoliang and Wang, Cheng and Qi, Jianzhong},
  booktitle={Proceedings of the AAAI conference on artificial intelligence},
  volume={34},
  number={01},
  pages={1234--1241},
  year={2020}
}

@inproceedings{liang2023airformer,
  title={Airformer: Predicting nationwide air quality in china with transformers},
  author={Liang, Yuxuan and Xia, Yutong and Ke, Songyu and Wang, Yiwei and Wen, Qingsong and Zhang, Junbo and Zheng, Yu and Zimmermann, Roger},
  booktitle={Proceedings of the AAAI conference on artificial intelligence},
  volume={37},
  number={12},
  pages={14329--14337},
  year={2023}
}

@article{jacobs1991adaptive,
  title={Adaptive mixtures of local experts},
  author={Jacobs, Robert A and Jordan, Michael I and Nowlan, Steven J and Hinton, Geoffrey E},
  journal={Neural computation},
  volume={3},
  number={1},
  pages={79--87},
  year={1991},
  publisher={MIT Press}
}

@inproceedings{sui2022causal,
  title={Causal attention for interpretable and generalizable graph classification},
  author={Sui, Yongduo and Wang, Xiang and Wu, Jiancan and Lin, Min and He, Xiangnan and Chua, Tat-Seng},
  booktitle={Proceedings of the 28th ACM SIGKDD conference on knowledge discovery and data mining},
  pages={1696--1705},
  year={2022}
}

@book{nesterov2004introductory,
  title={Introductory Lectures on Convex Optimization: A Basic Course},
  author={Nesterov, Yurii},
  year={2004},
  publisher={Springer}
}

\appendix

\section{Proofs and Full Statements}
\label{sec:prop}

\begin{proof}

After prototype alignment, each client-specific codebook can be decomposed
into the optimized global codebook $Q^*$ and a client-dependent environmental shift $\Delta_k^Q$:

\begin{equation}
\Pi_kQ_k
=
Q^*
+
\Delta_k^Q .
\label{eq:local-prototype-shift}
\end{equation}
Here, $\Delta_k^Q$ captures the deviation caused by the bias between the
local environmental distribution $P_k(E)$ and the global environmental
distribution $P(E)$.
A client observing only a subset of environmental regimes may therefore learn
a biased prototype codebook.

The federated aggregation combines aligned local codebooks according to the
aggregation weights $w_k$:

\begin{equation}
\begin{aligned}
Q =
\sum_k w_k\Pi_kQ_k =
\sum_k w_k
(Q^*+\Delta_k^Q).
\end{aligned}
\label{eq:federated-prototype-aggregation}
\end{equation}
Since the aggregation weights satisfy $\sum_k w_k=1$, we obtain
\begin{equation}
\begin{aligned}
Q
&=
Q^*
+
\sum_k w_k\Delta_k^Q .
\end{aligned}
\label{eq:global-prototype-shift}
\end{equation}
Therefore,
\begin{equation}
Q-Q^*
=
\sum_k w_k\Delta_k^Q ,
\end{equation}
which directly gives
\begin{equation}
\left\|
Q-Q^*
\right\|
=
\left\|
\sum_k w_k\Delta_k^Q
\right\|.
\label{eq:prototype-error}
\end{equation}

Next, we analyze how the prototype deviation propagates to the learned representation.
The prototype assignment operator is defined as
\begin{equation}
\hat z = A(z;Q)
=
\operatorname{softmax}
\left(
\frac{\cos(z,Q^\top)}{\alpha}
\right)Q . \notag
\end{equation}
By the continuity of cosine similarity and softmax, the assignment operator is locally Lipschitz continuous with respect to the prototype codebook~\citep{nesterov2004introductory}. Thus, there exists a constant $L_A>0$ such that
\begin{equation}
\left\|
A(z;Q)-A(z;Q^*)
\right\|
\leq
L_A
\left\|
Q-Q^*
\right\|.
\label{eq:assignment-bound}
\end{equation}

Substituting Eq.~\eqref{eq:prototype-error} into
Eq.~\eqref{eq:assignment-bound}, we obtain

\begin{equation}
\begin{aligned}
\left\|
\hat z
-
z^*
\right\|=
\left\|
A(z;Q)-A(z;Q^*)
\right\|
\leq
L_A
\left\|
\sum_k w_k\Delta_k^Q
\right\|.
\end{aligned}
\end{equation}
Absorbing the constant $L_A$ into the big-$O$ notation gives

\begin{equation}
\left\|
\hat z
-
z^{*}
\right\|
\leq
O
\left(
\left\|
\sum_k w_k\Delta_k^Q
\right\|
\right).
\end{equation}

Therefore, the federated de-confounding error (manifested as representation error) is
linearly controlled by the federation-averaged confounding effects (manifested as codebook deviations).
\end{proof}

\section{Hyperparameter Sensitivity}
\label{sec:app-param}
\begin{figure}[!h]
  \centering
  \includegraphics[width=0.9\linewidth]{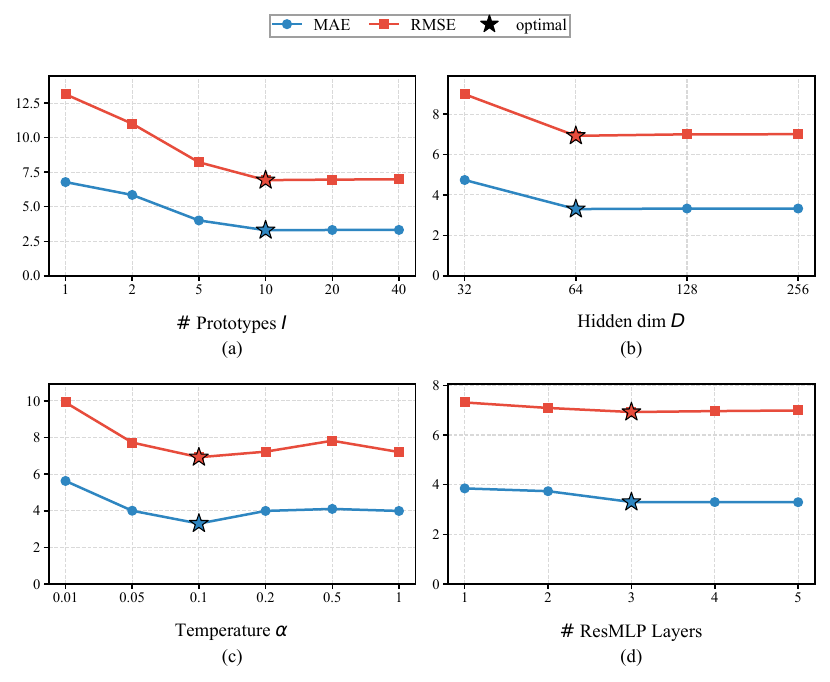}
  \caption{\textbf{Hyper-parameter sensitivity on \textsc{METR-LA}.}
We investigate the number of the prototypes $I$, hidden dimension $D$, assignment temperature $\alpha$, and encoder depth.
The starred markers denote the default configuration adopted throughout the paper ($I=10$, $D=64$, $\alpha=0.1$, and $3$ ResMLP layers).
}
  \label{fig:hyperparameters}
\end{figure}
Figure~\ref{fig:hyperparameters} shows that \method is generally robust to hyper-parameter choices.
Increasing the number of prototypes $I$ or hidden dimension $D$ consistently improves performance up to moderate values, after which both MAE and RMSE become nearly stable.
The assignment temperature $\alpha$ exhibits a clear optimum at $\alpha=0.1$, while both smaller and larger values lead to degraded performance. The results suggest that prototype assignment requires an appropriate balance between overly soft or sharp matching.
Finally, increasing the encoder depth beyond three ResMLP layers yields only marginal improvements, indicating that the performance gains mainly originate from the proposed de-confounding mechanism rather than a deeper backbone.

\end{document}